\documentclass[journal,twoside]{IEEEtran}

\usepackage{cite}
\usepackage{times}
\usepackage{epsfig}
\usepackage{graphicx}
\usepackage{amsmath}
\usepackage{amssymb}
\usepackage{acronyms}
\usepackage{subcaption}

\usepackage{hyperref}
\hypersetup{
    colorlinks=true,
    linkcolor=blue,
    filecolor=magenta,      
    urlcolor=red,
}

\DeclareMathOperator*{\argmax}{argmax}

\title{Point Set Voting for Partial Point Cloud Analysis}

\author{Junming Zhang$^{1}$, Weijia Chen$^{2}$, Yuping Wang$^{1}$, Ram Vasudevan$^{3}$ and Matthew Johnson-Roberson$^{4}$%

\thanks{This work was supported by a grant from Ford Motor Company via the Ford-UM Alliance under award  N028603.} 

\thanks{$^{1}$J. Zhang  and Y. Wang are with the Department of Electrical Engineering and Computer Science, University of Michigan, Ann Arbor, MI 48109 USA (email: junming, ypw@umich.edu)}
\thanks{ $^{2}$W. Chen is with the Robotics Program, University of Michigan, Ann Arbor, MI 48109 USA (email: weijiac@umich.edu)}
\thanks{$^{3}$ R. Vasudevan is with the Department of Mechanical Engineering, University of Michigan, Ann Arbor, MI 48109 USA ( email: ramv@umich.edu)}
\thanks{$^{4}$ M. Johnson-Roberson is with the Department of Naval Architecture and Marine Engineering, University of Michigan, Ann Arbor, MI 48109 USA (email: mattjr@umich.edu)}
}

\begin{document}

\maketitle

\begin{abstract}
The continual improvement of 3D sensors has driven the development of algorithms to perform point cloud analysis.
In fact, techniques for point cloud classification and segmentation have in recent years achieved incredible performance driven in part by leveraging large synthetic datasets. 
Unfortunately these same state-of-the-art approaches perform poorly when applied to incomplete point clouds. 
This limitation of existing algorithms is particularly concerning since point clouds generated by 3D sensors in the real world are usually incomplete due to perspective view or occlusion by other objects. 
This paper proposes a general model for partial point clouds analysis wherein the latent feature encoding a complete point cloud is inferred by applying a point set voting strategy.
In particular, each local point set constructs a vote that corresponds to a distribution in the latent space, and the optimal latent feature is the one with the highest probability. 
This approach ensures that any subsequent point cloud analysis is robust to partial observation while simultaneously guaranteeing that the proposed model is able to output multiple possible results. 
This paper illustrates that this proposed method achieves the state-of-the-art performance on shape classification, part segmentation and point cloud completion. The code is available at \url{https://github.com/junming259/PointSetVoting}.

\end{abstract}

\section{Introduction}

The quality of 3D sensors has rapidly improved in recent years as their ability to accurately and quickly measure the depth of scenes has surpassed vision-based methods~\cite{vzbontar2016stereo, kendall2017end, zhang2019dispsegnet}. 
This improved accessibility to point clouds demands the development of algorithms to interpret and analyze them. 
Inspired by the success of \acp{DNN} in solving 2D image analysis tasks, approaches with \acp{DNN} have been successfully applied to perform similar point cloud analysis tasks such as shape classification and part segmentation~\cite{qi2017pointnet++,li2018pointcnn,qi2017pointnet,liu2019relation,wu2019pointconv,wang2019dynamic,thomas2019kpconv}. 
These \acp{DNN} methods achieve the state-of-the-art performance on these point cloud analysis tasks by learning representations from large synthetic datasets constructed from sampling the surfaces of CAD objects~\cite{wu20153d,chang2015shapenet}.
Unfortunately since point clouds generated from 3D sensors in the real-world scenarios are often incomplete, these approaches struggle when tasked to perform shape classification or part segmentation on partial point clouds. 
Since real-world point clouds are often incomplete due to perspective of view or occlusions, this shortcoming of existing point cloud analysis methods is particularly cumbersome.

To address the limitations of existing approaches, this paper proposes a novel model for partial point clouds analysis.
We model the point clouds as a partition of point sets. 
Each local point set independently contributes to infer the latent feature encoding the complete point cloud. 
In contrast to prior work on learning a representation of all points within the point clouds, we utilize an encoder to embed each local point set.
All embeddings vote to infer a latent space characterized by a distribution. 
As we show in this paper, giving each local point set a vote ensures that the model has the ability to address the incomplete nature of point clouds.
Inspired by recent progress in variational inference~\cite{diederik2014auto,kingma2019introduction}, we output a distribution for the latent variable and then use a decoder to generate a prediction from the latent value with the highest probability. 
In particular, each local point set generates a Gaussian distribution in the latent space and independently votes to form the distribution of the latent variable.
This voting strategy ensures that the model outputs more accurate prediction when more partial observations are given, and the probabilistic modeling enables the model to generate multiple possible outputs.  

The contributions of this paper are: 
(1) We propose that each local point set independently votes to infer the latent feature. This voting strategy is shown to be robust to partial observation;
(2) We propose to construct each vote as a distribution in the latent space and this distribution modeling allows for diverse predictions;
(3) The proposed model trained with complete point clouds using the proposed training strategy performs robustly on partial observation at test, which reduces the cost of collecting large partial point clouds dataset;
(4) The proposed model achieves state-of-the-art results on shape classification, part segmentation, and point clouds completion. In particular, it outperforms approaches trained with pairs of partial and complete point clouds on point clouds completion.

\section{Related Work}

To perform point cloud analysis, researchers have traditionally converted point clouds into 2D grids or 3D voxels since they can leverage existing \acp{CNN}. 
With the help of \acp{CNN}, those approaches achieve impressive results on 3D shape analysis~\cite{feng2018gvcnn,voxnet,qi2016volumetric}.
Unfortunately, these 2D grid or 3D voxel representations degrade the resolution of objects. 
Researchers have attempted to address this issue by utilizing sparse representations~\cite{tatarchenko2017octree,wang2017cnn}.
However, these representations are still less efficient than point clouds and are unable to avoid quantization effects. 
More recently, PointNet~\cite{qi2017pointnet} has pioneered the approaches of directly taking point clouds as inputs and processed them using \acp{DNN}. 
To accomplish this objective, it uses a symmetric function to aggregate information from each point which is transformed to a high-dimensional space. A variety of extensions have been applied to PointNet \cite{qi2017pointnet++,liu2019relation,wang2019dynamic,shen2018mining}; however, none of them is robust to the partial observation that is common in real-world scenarios.

To address this challenge posed by partial observations,
researchers have relied on training \ac{DNN}s on partial point clouds collected in real-world scenarios~\cite{qi2019deep,shi2019pointrcnn,qi2018frustum,dai2017scannet}.
Each of these approaches rely on networks that are proposed to perform feature extraction on complete point clouds. 
Unfortunately, collecting and annotating those partial point cloud datasets are expensive.
Another approach seeks to first infer the complete data of the partially observed point clouds before later analysis.
A common pipeline to perform this completion first encodes partial observations into a feature vector and then decodes it to complete point clouds. 
A variety of methods have been proposed for designing the  decoder~\cite{fan2017point,sun2020pointgrow,yang2018foldingnet,yuan2018pcn,tchapmi2019topnet,Stutz_2018_CVPR}. 
However, each of these methods outputs a single prediction given partial shapes and lacks the ability to generate multiple plausible results. 
One notable exception is able to generate diverse results by modeling the spatial distribution of all points~\cite{sun2020pointgrow}. 
Unfortunately, this approach is only able to address partial point clouds from specific locations.
 In contrast, the method developed in this paper has no such requirement on an observed partial point cloud, and leverages the distribution over the latent space to generate diverse predictions.

Hough transform and its variations have been applied to solve many problems, such as pattern detection, object detection and pose estimation~\cite{duda1972use,ballard1981generalizing,borrmann20113d,sun2010depth,kehl2016deep,guo2008probabilistic}.
Recent work, VoteNet~\cite{qi2019deep}, extends it to 3D object detection and demonstrates that hough voting is well suited for 3D point clouds analysis.
Specifically, each local point set votes for the center of the object, and only those votes which locate within the cluster are considered for later process.
Similarly, we take advantage of this voting strategy to accumulate small bits of partial information to form a confident latent feature.
However, we model each vote as a distribution in the latent space instead of a deterministic feature vector. 
This probabilistic modeling enables the model to learn the weights of votes when predicting latent features and is also useful to generate multiple possible outputs if needed.

The \ac{VAE} is one of the popular methods to model a generative distribution~\cite{diederik2014auto}. 
It assumes a prior distribution of latent variables, which is often a Gaussian distribution. 
More recently, the \ac{CVAE} was proposed to extend the \ac{VAE} by modeling the conditional distribution~\cite{sohn2015learning}.
Unfortunately, directly applying a \ac{CVAE} to partial point clouds requires a collection of annotated partial point cloud datasets for training. This paper addresses this limitation by proposing each local point set serve as the unit voter to contribute to the latent feature. Encoding features learned for local point sets of complete point clouds can be leveraged for embedding local point sets of partial point clouds at test, which allows us to train on complete point clouds and perform on partial point clouds at test. 




\section{Problem Statement}
Consider an observed partial point cloud, denoted by $\mathbf{x} \subseteq {\rm I\!R}^3$.
Suppose the output of a model for a point cloud analysis task is $\mathbf{y}$ and we are interested in modeling the conditional distribution $p(\mathbf{y}|\mathbf{x})$, which provides a way to predict $\mathbf{y}$ given the partial inputs $\mathbf{x}$. 
This paper aims to address the following three challenges: (1) classifying partial point clouds, (2) segmenting parts on partial point clouds, and (3) recovering complete point clouds from partial observation.

\section{Method}

This section describes the method we use to accomplish the aforementioned objective.

\begin{figure}[t]
    \centering
    \includegraphics[width=\linewidth]{{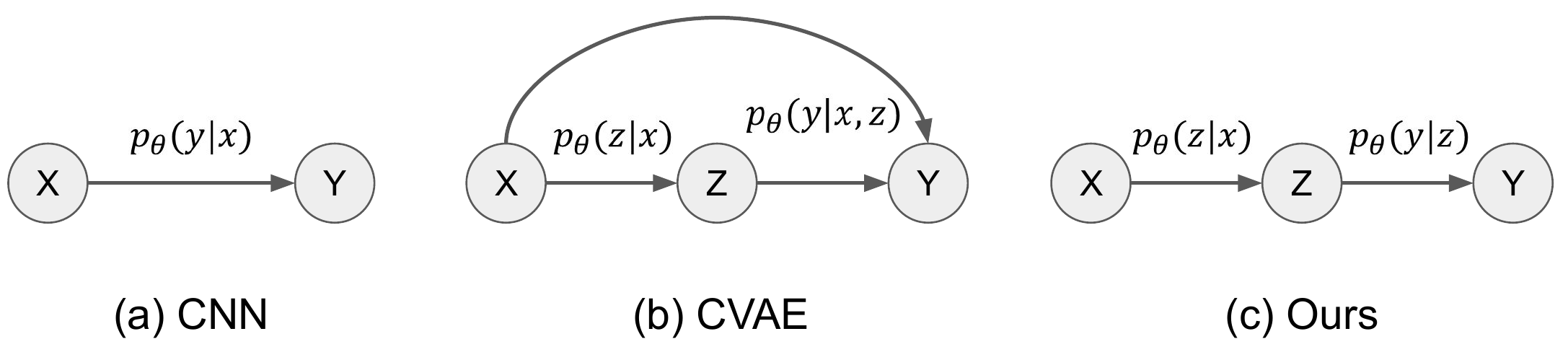}}
    \caption{An illustrative comparison between our proposed method and other conditional graphical models (CGMs).}
    \label{fig:graph_model}
\end{figure}

\begin{figure*}[t!]
    \centering
    \includegraphics[width=\textwidth]{{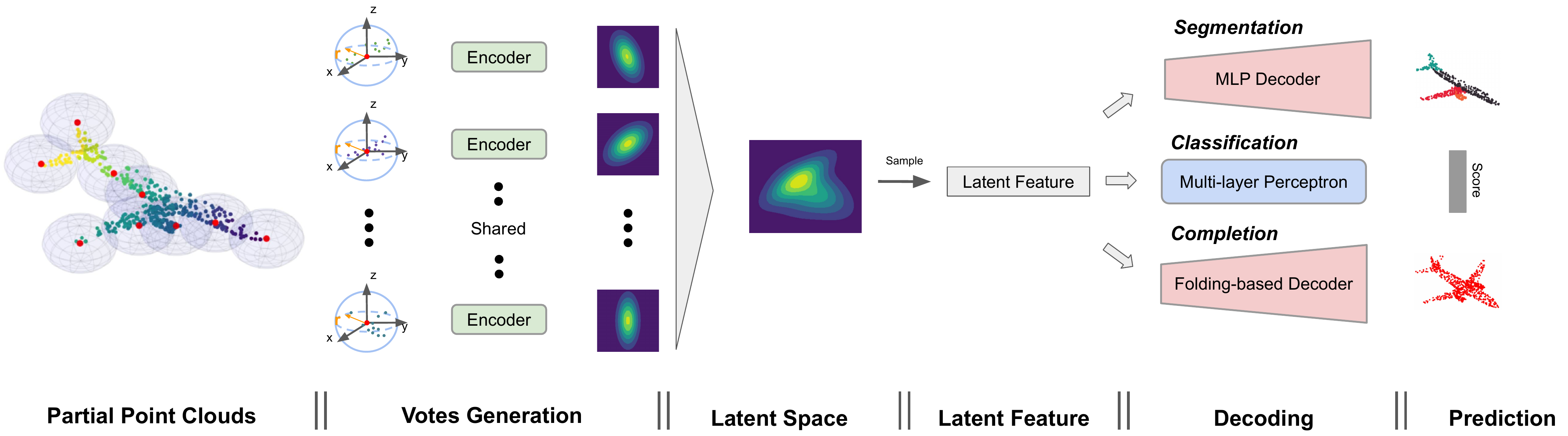}}
    \caption{An illustration of the model developed in this paper. The input point clouds are modeled as a partition of point sets and each local point set is defined by its centroid and scale $r$. Local point sets are embedded by a shared-weight PointNet encoder. The voting via this encoder is used to infer the space for latent features. The latent feature with the highest probability is sampled from the inferred latent space and is then passed to a decoding module for prediction. We perform experiments on three tasks: classification, part segmentation and point clouds completion.}
    \label{fig:architecture}
\end{figure*}

\subsection{Preliminary: Conditional Variational Auto-encoder}
A \ac{CVAE}~\cite{sohn2015learning} is a directed graphical model.
The conditional generative process of CVAE is as follows: for a given observation $\mathbf{x}$, a noise vector $\mathbf{z}$ is drawn from the prior distribution $p_{\theta}(\mathbf{z} | \mathbf{x})$, and an output $\mathbf{y}$ is generated from the distribution $p_{\theta}(\mathbf{y} | \mathbf{x}; \mathbf{z})$. 
The training objective, variational lower bound, of \ac{CVAE} is written as follows:

\begin{equation}
\begin{aligned}
    \mathrm{log} \: p_{\theta}(\mathbf{y}|\mathbf{x}) \geq \mathcal{L}_{\phi, \theta}(\mathbf{y}|\mathbf{x}) =  -\mathbf{KL}(q_{\phi}(\mathbf{z}|\mathbf{x}, \mathbf{y})||p_{\theta}(\mathbf{z}|\mathbf{x}))) \\ + E_{q_{\phi}(\mathbf{z}|\mathbf{x}, \mathbf{y})}[\mathrm{log} \: p_{\theta}(\mathbf{y}|\mathbf{x}, \mathbf{z})]
\end{aligned}
\end{equation}

The \ac{CVAE} is composed of recognition network $q_{\phi}(\mathbf{z}|\mathbf{x}; \mathbf{y})$, prior network $p_{\theta}(\mathbf{z}|\mathbf{x})$, and generation network
$p_{\theta}(\mathbf{y}|\mathbf{x}; \mathbf{z})$. 
In this framework, the recognition network $q_{\phi}(\mathbf{z}|\mathbf{x}; \mathbf{y})$ is used to approximate the prior network  $p_{\theta}(\mathbf{z}|\mathbf{x})$. 
All distributions are modeled using neural networks. 
During training, the reparameterization trick~\cite{diederik2014auto} is applied to propagate the gradients of $\phi$ and $\theta$ through the latent variables $\mathbf{z}$. 

\subsection{Proposed Point Cloud Model}

We model the point clouds $\mathbf{x}$ as an overlapping partition of point sets, denoted by $\{x_1, x_2, ..., x_n\}$ if there are $n$ point sets in the partition.
In the simplest setting, each point is described by just its 3D coordinates, i.e. $x_i \subseteq {\rm I\!R}^3$.
Each point set is defined by a centroid and scale, as is shown in the Figure \ref{fig:architecture}. 
To evenly cover the whole point clouds, we use Farthest Point Sampling (FPS)~\cite{qi2017pointnet++} algorithm to sample centroids. 
The number of centroids and the scale are manually set.


\subsection{Proposed Method}

In contrast to \ac{CVAE}s, in which the generation network takes $(\mathbf{x}, \mathbf{z})$ as inputs, we model the generation process as a Markov chain, as is shown in the Figure \ref{fig:graph_model}. 
Specifically, given the latent variable $\mathbf{z}$ sampled from $p(\mathbf{z}|\mathbf{x})$, $\mathbf{y}$ is independent on $\mathbf{x}$. 
As a result, the generation of the output $\mathbf{y}$ satisfies the following equation: $p(\mathbf{y}|\mathbf{x},\mathbf{z}) = p(\mathbf{y}|\mathbf{z})$.
The variational lower bound of this model is written as follows:

\begin{equation}
\label{eq:vlb_ours}
\begin{aligned}
    \mathcal{L}_{\phi, \theta}(\mathbf{y}|\mathbf{x}) =  -\mathbf{KL}(q_{\phi}(\mathbf{z}|\mathbf{x})||p_{\theta}(\mathbf{z}|\mathbf{x}))) \\ + E_{q_{\phi}(\mathbf{z}|\mathbf{x})}[\mathrm{log} \: p_{\theta}(\mathbf{y}|\mathbf{z})]
\end{aligned}
\end{equation}

One problem with this learning framework is that the generation network $p_{\theta}(\mathbf{y}|\mathbf{z})$ takes values sampled from the recognition network $q_{\phi}(\mathbf{z}|\mathbf{x})$ at training while takes values sampled from the prior network $p_{\theta}(\mathbf{z}|\mathbf{x})$ at testing.
This makes training inconsistent with testing. 
Similar to~\cite{sohn2015learning}, we force consistency between these settings by making the recognition network $q_{\phi}(\mathbf{z}|\mathbf{x})$ the same as the prior network $p_{\theta}(\mathbf{z}|\mathbf{x})$.
By doing this, $\mathbf{z}$ can be drawn from the distribution $q_{\phi}(\mathbf{z} | \mathbf{x})$ at both training and testing, and the KL divergence term becomes zero. 
We approximate the resulting version of Equation \eqref{eq:vlb_ours} with the Monte Carlo estimator formed by sampling $\mathbf{z}^{(i)}$ from the recognition network $q_{\phi}(\mathbf{z} | \mathbf{x})$:
\begin{equation}
    \mathcal{L}_{\phi, \theta}(\mathbf{y}|\mathbf{x}) \approx \frac{1}{L} \sum_{i=1}^{L}\mathrm{log} \: p_{\theta}(\mathbf{y}|\mathbf{z}^{(i)}), \; \mathbf{z}^{(i)} \sim q_{\phi}(\mathbf{z}|\mathbf{x}).
\label{eq:monte_carlo}
\end{equation}

Recall that in our case $\mathbf{x}$ is modeled as a set of local point sets $\{x_i\}_{i=1}^n$.
Thus, we propose to generate a vote for each of them, all of which are used to compute the latent variable $\mathbf{z}$. This voting strategy is inspired by the Hough transform~\cite{duda1972use} and VoteNet~\cite{qi2019deep}, and it is shown to accumulate small bits of partial information and output confident predictions.
We model each vote as a distribution in the latent space. 
By assuming the independence of each vote, the recognition network $q_{\phi}(\mathbf{z}|\mathbf{x})$ can be expanded as follows:
\begin{equation}
    q_{\phi}(\mathbf{z}|\mathbf{x}) = \prod_{i=1}^{n}q_{\phi}(\mathbf{z}|x_i).
\end{equation}

In the experiments, we assume $q_{\phi}(\mathbf{z}|x_i)$ is a Gaussian distribution characterized by mean vector and covariance matrix, which enables the use of a closed-form solution to optimizing $q_{\phi}(\mathbf{z}|\mathbf{x})$ with respect to $\mathbf{z}$. 
We denote the maximizing argument of $q_{\phi}(\mathbf{z}|\mathbf{x})$ by $\mathbf{z}_{opt}$.
Equation~\eqref{eq:monte_carlo} is equivalent to estimation with a single point when setting $L=1$. 
Combining this with the highest probability sample of $\mathbf{z}$, given by $\mathbf{z}_{opt}$, the objective function can be further written as follows:
\begin{equation}
\begin{aligned}
    \mathcal{L}_{\phi, \theta}(\mathbf{y}|\mathbf{x}) \approx \mathrm{log} \: p_{\theta}(\mathbf{y}|\mathbf{z}_{opt}), \quad \mathbf{z}_{opt} = \argmax_z \: \prod_{i=1}^{n}q_{\phi}(\mathbf{z}|x_i).
\end{aligned}
\end{equation}

Previous variational models choose latent features by sampling.
However, $\mathbf{z}_{opt}$ in our case can be computed directly, so all operations are differentiated with respect to the parameters $\theta$ and $\phi$, which means that the reparameterization trick is no longer needed. 

Note that the loss function differs as the task changes. 
A common softmax cross-entropy loss is used for classification and segmentation whereas  the Chamfer distance~\cite{fan2017point} is used for training models on point cloud completion task.
After training, the generative process is as follows: for the given observation $\mathbf{x}$, $\mathbf{z}_{opt}$ is the result of voting from $\{q_{\phi}(\mathbf{z} | x_i)\}_{i=1}^n$, and then the output $\mathbf{y}$ is generated from $p_{\theta}(\mathbf{y} | \mathbf{z}_{opt})$. 
The use of $\mathbf{z}_{opt}$ produces a single deterministic prediction.
Diverse predictions are generated by instead sampling $\mathbf{z}^{(i)} \! \sim \! q_{\phi}(\mathbf{z}|\mathbf{x})$ followed by applying the generation network as in the deterministic case.

\section{Experiment}

\subsection{Implementation}
\label{sec:implementation}

\textbf{Network architecture}\; The architecture of the proposed model is illustrated in the Figure \ref{fig:architecture}. 
We use \acp{DNN} to model $q_{\phi}(\mathbf{z}|\mathbf{x})$ and $p_{\theta}(\mathbf{y}|\mathbf{z})$. 
A shared-weights network is used to represent $q_{\phi}(\mathbf{z}|x_i)$. 
Given the local point set, we represent each point within the local region relative to the centroid and then use a shared-weight PointNet as the basic feature extractor to encode the local region.
Both the encoding feature and coordinates of centroid are processed by a \ac{MLP} and it outputs a distribution of the latent space.
We assume a simple case of multivariate Gaussian distribution with diagonal covariance matrix. 
So the output from the \ac{MLP} consists of two vectors representing the mean and diagonal elements in the covariance matrix respectively.
Different downstream tasks correspond to different networks for modeling $p_{\theta}(\mathbf{y}|\mathbf{z})$ as is shown in Figure \ref{fig:architecture}.
A folding-based decoder~\cite{yang2018foldingnet} is used for point clouds completion. 

\textbf{Training Strategy}\; During training, we partition the point cloud into 64 local point sets, and each of them generates a vote in the latent space. 
To make the voting strategy tenable, we propose to drop some votes and only a small portion of votes contribute to infer the latent feature at training.
We do random selection and in extreme case only one vote is selected. 
This ensures that a single vote still has the potential to be decoded to a reasonable prediction. 
Thus, the learned embedding can be leveraged by the basic unit voter at test and improves the robustness to any type partial shapes. 
Note that all votes contribute to compute latent feature at testing.

\begin{figure}
\centering
  \includegraphics[width=\linewidth]{{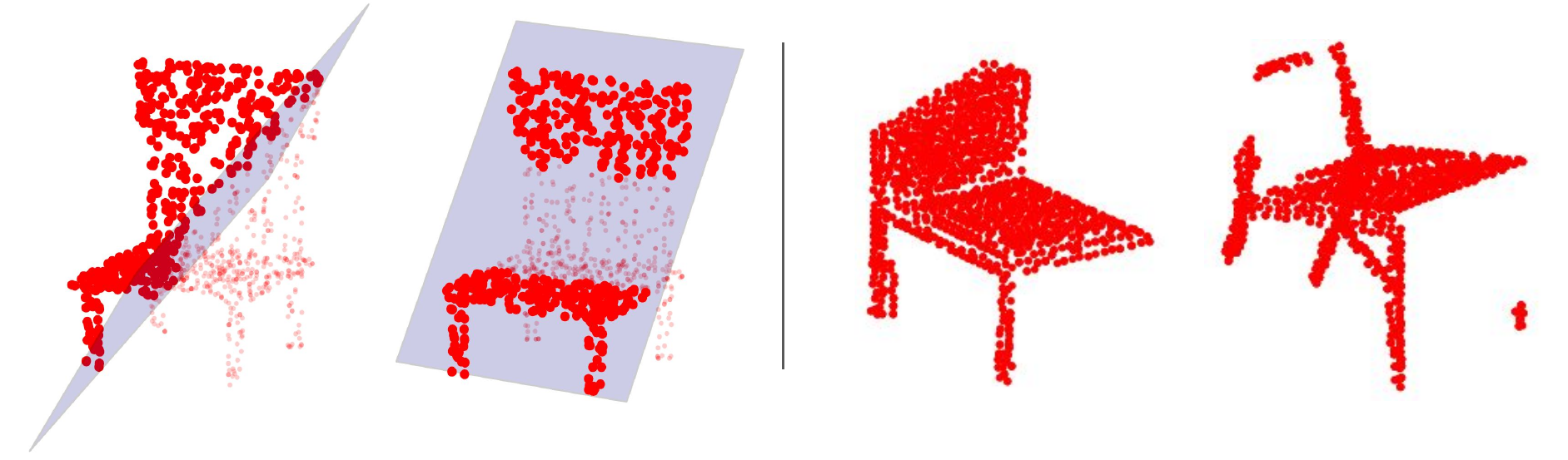}}
  \caption{Sample partial point clouds. Left: partial point clouds synthesized by choosing points (bolded red) falling into one side of a random 2D plane. Right: partial point clouds generated by back-projecting depth images into 3D space.}
  \label{fig:partial_point_clouds}
\vspace{-5mm}
\end{figure}

\textbf{Partial Point clouds}\; 
Partial point clouds are only used at testing.
In this work, we experiment on two kinds of partial point clouds.
Sample partial point clouds are visualized in the Figure \ref{fig:partial_point_clouds}.
For point cloud datasets that do not provide partial point clouds, we adapt a simple strategy to simulate partial point clouds, which are synthesized by selecting points falling into one side of a plane. 
The plane goes through the origin and is defined by the normal, which is a 3D vector and generated by randomly sampling from a normal distribution.
The other partial point clouds used in this work are generated by back-projecting depth images into 3D space, and they are provided by Completion3D dataset~\cite{tchapmi2019topnet}. 
Compared to simulating partial point clouds using a subset of complete point clouds, these generated partial point clouds are closer to real-world sensor data.

\subsection{Point Clouds Classification}
\label{sec:point_clouds_classification}

We  first consider the task of point cloud classification. 
This requires that the model extract global features describing distinct geometric information and decode it into a predicated category.

\textbf{Dataset}\; We use the ModelNet40~\cite{wu20153d} dataset to evaluate our proposed method on shape classification of point clouds. 
It contains 12,311 shapes from 40 object categories. 
In the experiments, point clouds are generated by evenly sampling 1024 points from the surface of objects.
We follow the same strategy of set splitting as in~\cite{qi2017pointnet}. 
Before being passed into the model, point clouds are centered and normalized within a unit sphere. No data augmentation techniques are applied during training.

\textbf{Results}\; Quantitative results on Modelnet40 are shown in the Table \ref{table:modelnet40}. 
Overall classification accuracy is reported on both complete point clouds in the column (Complete) and simulated partial point clouds in the column (Partial). 
All listed methods are trained on provided training set and achieve the state-of-the-art results on complete point clouds.
Our proposed method performs slightly better than PointNet and PointNet++.
When evaluated on simulated partial point clouds, however, other approaches experience a significant drop in accuracy. 
This is unsurprising since existing approaches are designed and trained for complete point clouds, and generalize poorly to novel partial point clouds. 
In contrast, our method trained on complete point clouds using the proposed training strategy is robust to partial observation and achieves 86.4\% classification accuracy on partial point clouds. Analysis of the proposed training strategy will be discussed in the later section. 

\begin{table}[t]
\begin{center}
\begin{tabular}{l|cccccc}
\hline
Method & Input & Complete & Partial & Partial*\\
\hline
PointNet~\cite{qi2017pointnet} & $xyz$ & 88.8 & 20.9 & 76.5 \\
PointNet++~\cite{qi2017pointnet++} & $xyz$ & 91.0 & 61.5 & 81.9 \\
RS-CNN~\cite{liu2019relation} & $xyz$ & 92.3 & 43.3 & 71.1\\
DG-CNN~\cite{wang2019dynamic} & $xyz$ & \textbf{92.9} & 51.5 & 64.9\\
\hline
Ours & $xyz$ & 91.4 & \textbf{86.4} & \textbf{86.4} \\
\hline
\end{tabular}
\end{center}
\caption{Classification accuracy on ModelNet40. Overall classification accuracy is reported on complete point clouds (Complete) and simulated partial point clouds (Partial). The last column (Partial*) shows the results when models trained using the proposed training strategy}
\label{table:modelnet40}
\vspace{-5mm}
\end{table}

\begin{figure*}[t]
    \centering
    \includegraphics[width=0.8\linewidth]{{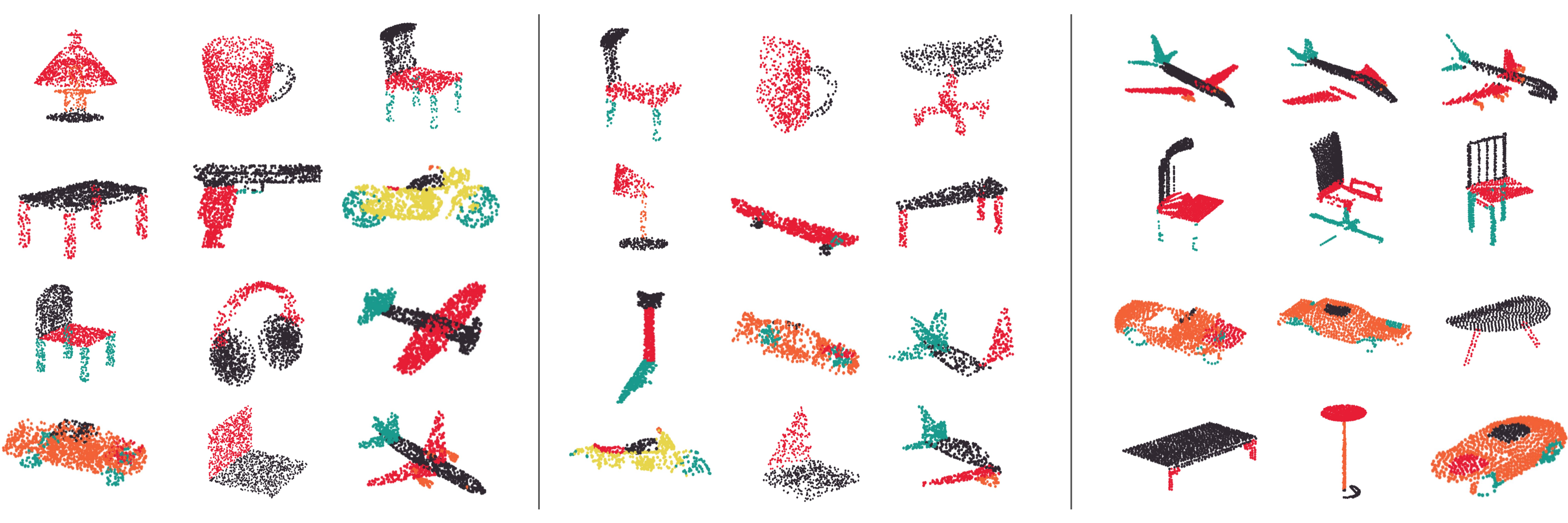}}
    \caption{Qualitative results on part segmentation of point clouds using the method presented in this paper.  Different colors correspond to distinct segments. Left: segmentation on the ShapeNet part dataset. Middle: segmentation on the simulated partial point clouds in the ShapeNet. Right: segmentation on the Completion3D dataset.}
    \label{fig:segmentation}
\end{figure*}

\subsection{Part Segmentation}
Given the point clouds and object category, the part segmentation tasks requires predicting a part label for each point. 
As a result, this task requires that the model extract both global and local information. 

\textbf{Dataset}\; We use the ShapeNet part dataset~\cite{wu20153d} to evaluate our proposed method on part segmentation of point clouds. 
It contains 16,881 shapes from 16 object categories with 50 parts. 
Each point cloud contains 2048 points which are generated by evenly sampling from the surface of objects. 
We follow the same set splitting conventions in~\cite{qi2017pointnet++}. 
During evaluation, mean inter-over-union (mIoU) that are averaged across all classes is reported. 
No data augmentation techniques are applied during training.

\textbf{Results}\; We report the results of our proposed method and compare its performance to existing approaches in the Table \ref{table:shapenet}. 
All methods are trained on provided training set.
Our proposed method has superior performance on partial point clouds and achieves 78.1 mIoU, while others achieve around 30 mIoU. 
This robustness to partial observation comes at the expense of slightly lower accuracy on segmentation of complete point clouds when compared with other approaches. 
We also test the robustness of our approach by applying it to the completion3D dataset~\cite{tchapmi2019topnet} which contains partial point clouds but no part labels. 
Qualitative results on part segmentation by applying the method proposed in this paper are shown in the Figure \ref{fig:segmentation}.

\begin{table}[t]
\begin{center}
\begin{tabular}{l|ccccc}
\hline
Method & Input & Complete & Partial \\
\hline
PointNet~\cite{qi2017pointnet} & $xyz$ & 80.5 & 29.9 \\
PointNet++~\cite{qi2017pointnet++} & $xyz$ & 82.0 & 30.9 \\
DG-CNN~\cite{wang2019dynamic} & $xyz$ & 82.3 & 29.8 \\
RS-CNN~\cite{liu2019relation} & $xyz$ & \textbf{82.4} & 30.6 \\
\hline
Ours & $xyz$ & 79.0 & \textbf{78.1} \\
\hline
\end{tabular}
\end{center}
\caption{Part segmentation results on ShapeNet part dataset. Mean intersection of unions (mIoUs) is reported on complete point clouds (Complete) and simulated partial point clouds (Partial). }
\label{table:shapenet}
\end{table}

\begin{table}[t]
\centering
\begin{tabular}{c|cccccc|c}
Model      &  BN     & DP      & \# v. train  & \# v. test & radius  & BK   & Acc.\\
\hline\hline
$A_1$           &            &            & 10         & 16         & 0.25      & 1024      & 78.4 \\
$A_2$           & \checkmark &            & 10         & 16         & 0.25      & 1024      & 80.9 \\
$A_3$           & \checkmark & \checkmark & 10         & 16         & 0.25      & 1024      & \textbf{81.0} \\
\hline
$B_1$           & \checkmark & \checkmark & 4          & 16         & 0.25      & 1024      & 79.1 \\
$B_2$           & \checkmark & \checkmark & 10         & 16         & 0.25      & 1024      & \textbf{81.0} \\
$B_3$           & \checkmark & \checkmark & 16         & 16         & 0.25      & 1024      & 79.4 \\
$B_4$           & \checkmark & \checkmark & 32         & 16         & 0.25      & 1024      & 78.1 \\
$B_5$           & \checkmark & \checkmark & 64         & 16         & 0.25      & 1024      & 76.2 \\
\hline
$C_1$           & \checkmark & \checkmark & 10         & 8          & 0.25      & 1024      & 76.8 \\
$C_2$           & \checkmark & \checkmark & 10         & 16         & 0.25      & 1024      & 81.0 \\
$C_3$           & \checkmark & \checkmark & 10         & 32         & 0.25      & 1024      & 82.7 \\
$C_4$           & \checkmark & \checkmark & 10         & 64         & 0.25      & 1024      & 83.8 \\
$C_5$           & \checkmark & \checkmark & 10         & 128        & 0.25      & 1024      & 84.4 \\
$C_6$           & \checkmark & \checkmark & 10         & 256        & 0.25      & 1024      & \textbf{84.6} \\
\hline
$D_1$           & \checkmark & \checkmark & 10         & 256        & 0.15      & 1024      & 83.1 \\
$D_2$           & \checkmark & \checkmark & 10         & 256        & 0.20      & 1024      & \textbf{86.4} \\
$D_3$           & \checkmark & \checkmark & 10         & 256        & 0.25      & 1024      & 84.6 \\
$D_4$           & \checkmark & \checkmark & 10         & 256        & 0.35      & 1024      & 82.5 \\
\hline
$E_1$           & \checkmark & \checkmark & 10         & 256        & 0.20      & 512       & 85.3 \\
$E_2$           & \checkmark & \checkmark & 10         & 256        & 0.20      & 1024      & 86.4 \\
$E_3$           & \checkmark & \checkmark & 10         & 256        & 0.20      & 2048      & \textbf{86.8} \\

\hline
\end{tabular}
\caption{Ablation study. The metric is overall classification accuracy on the simulated partial ModelNet40 test set. ``BN" indicates using batch normalization; ``DP" indicates using the dropout technique in fully connected layers except the final one; ``\# v. train" indicates the maximum number of votes selected to contribute to the latent feature at training; ``\# v. test" indicates the number of votes at test; ``radius'' indicates the ball radius of local regions; ``BK" indicates the dimension (bottleneck) of the latent space.}
\label{table:ablation_study}
\vspace{-3mm}
\end{table}

\subsection{Point Clouds Completion}
Given partial point clouds, point cloud completion tries to generate points to recover the complete shapes. 
This requires a model that has the ability to infer a global feature which encodes complete point clouds from the partial inputs.  

\textbf{Dataset} We evaluate our model on Completion3D~\cite{tchapmi2019topnet}, which is a 3D object point cloud completion benchmark. 
It contains pairs of partial and complete point clouds from 8 categories which are derived from the Shapenet dataset with 2048 points per object point clouds. 
We apply the set splitting given by the dataset. 
Partial point clouds are generated by back-projecting 2.5D depth images into 3D space and complete point clouds are used as ground truth.

\textbf{Results} Quantitative results on Completion3D's withheld test dataset are shown in Table \ref{table:completion3D} where the Chamfer distance multiplied by $10^4$ is reported. 
Our proposed model achieves 18.18 average Chamfer distance and outperforms FoldingNet (19.07) and PCN (18.22), which are trained with both partial and complete point clouds, while only complete point clouds are used during training for the method developed in this paper.

\begin{figure*}
    \centering
    \includegraphics[width=0.75\linewidth,height=0.2\linewidth]{{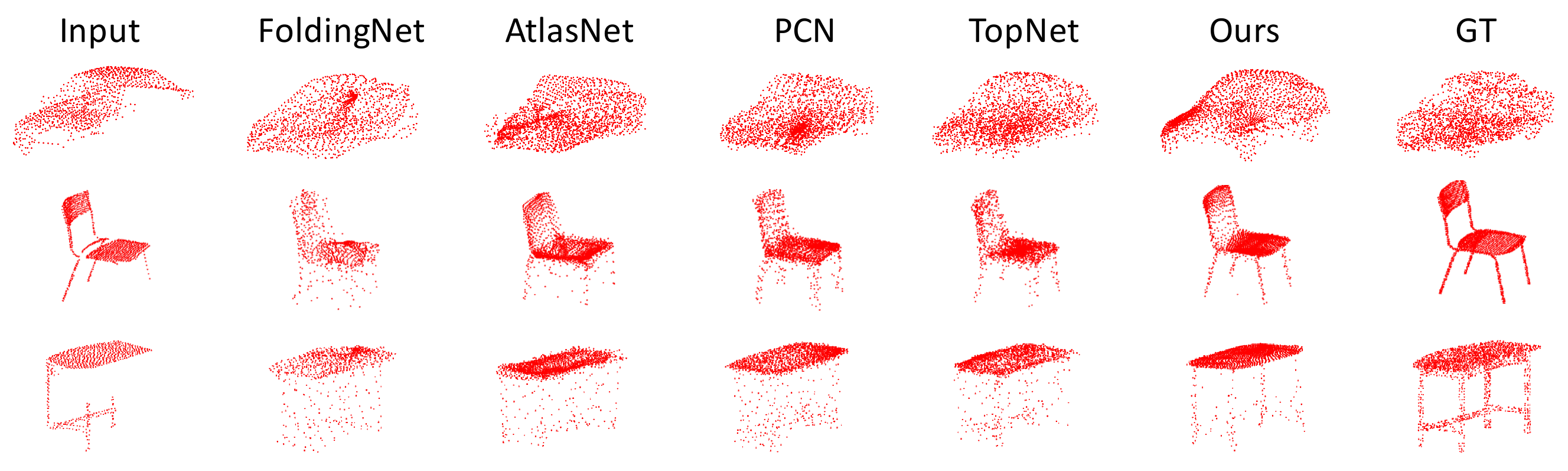}}
    \caption{Qualitative point clouds completion results on partial point clouds provided in Completion3D dataset.}
    \label{fig:completion}
\vspace{-3mm}
\end{figure*}

\begin{figure*}
    \centering
    \includegraphics[width=0.75\linewidth,height=0.2\linewidth]{{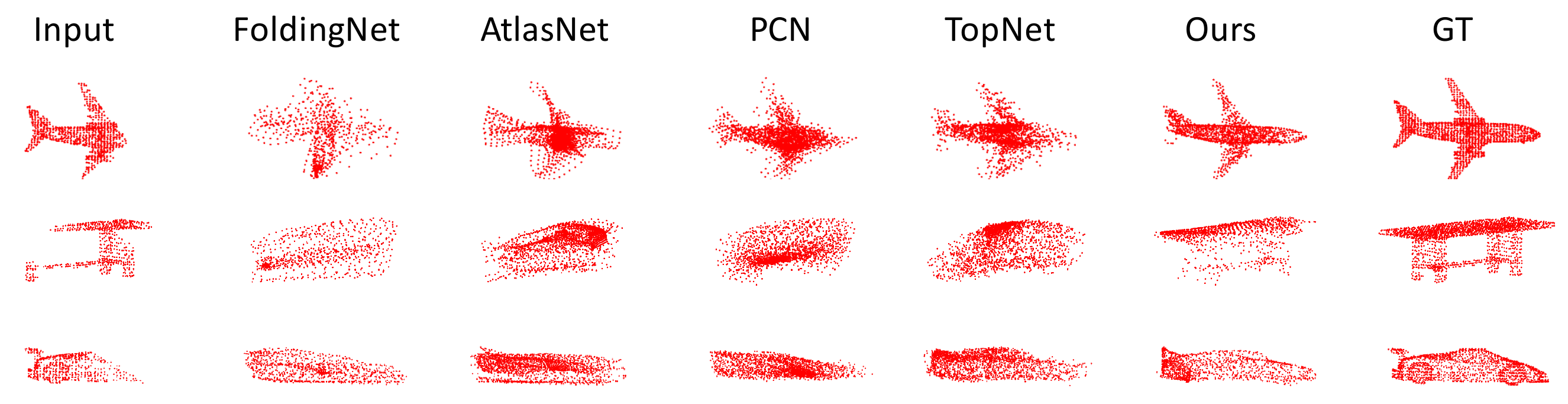}}
    \caption{Qualitative point clouds completion results on partial point clouds simulated on Completion3D dataset using the introduced strategy in the paper.}
    \label{fig:completion_simulated}
\vspace{-3mm}
\end{figure*}

\begin{table*}[t!]
\centering
    \begin{tabular}{l|cccccccc|c}
    Model & Plane & Cabinet & Car & Chair & Lamp & Sofa & Table & W.craft & Average \\
    \hline\hline
    FoldingNet~\cite{yang2018foldingnet} & 12.83 & 23.01 & 14.88 & 25.69 & 21.79 & 21.31 & 20.71 & 11.51 & 19.07 \\ 
    PCN~\cite{yuan2018pcn} & 9.79 & 22.70 & 12.43 & 25.14 & 22.72 & 20.26 & 20.27 & 11.73 & 18.22 \\ 
    AtlasNet~\cite{groueix2018papier} & 10.36 & 23.40 & 13.40 & 24.16 & 20.24 & 20.82 & 17.52 & 11.62 & 17.77 \\ 
    TopNet~\cite{tchapmi2019topnet} & 7.32 & \textbf{18.77} & \textbf{12.88} & \textbf{19.82} & \textbf{14.60} & \textbf{16.29} & \textbf{14.89} & \textbf{8.82} & \textbf{14.25}\\
    \hline
    Ours & \textbf{6.88} & 21.18 & 15.78 & 22.54 & 18.78 & 28.39 & 19.96 & 11.16 & 18.18 \\
    \hline
    \end{tabular}
    \caption{
    The performance of various state-of-the-art algorithms on partial point cloud completion on the Completion3D benchmark dataset. 
    Results are reported on the Completion3D's withheld test set. 
    The Chamfer distance (CD) is reported, multiplied by $10^4$.}
    \label{table:completion3D}
\vspace{-3mm}
\end{table*}    
    
\begin{table*}[t!]
\centering
    \begin{tabular}{l|cccccccc|c}
    Model & Plane & Cabinet & Car & Chair & Lamp & Sofa & Table & W.craft & Average \\
    \hline\hline
    FoldingNet~\cite{yang2018foldingnet} & 25.79 & 40.52 & 16.12 & 39.90 & 43.01 & 43.76 & 40.88 & 26.54 & 34.56 \\ 
    PCN~\cite{yuan2018pcn} & 21.58 & 41.87 & 16.56 & 38.86 & 50.19 & 43.37 & 39.44 & 27.57 & 34.93 \\ 
    AtlasNet~\cite{groueix2018papier} & 23.13 & 49.60 & 16.80 & 43.34 & 60.83 & 48.21 & 41.94 & 33.96 & 39.73 \\ 
    TopNet~\cite{tchapmi2019topnet} & 21.61 & \textbf{15.24} & 38.14 & 35.23 & 44.42 & 38.36 & 36.18 & 25.97 & 31.87\\
    \hline
    Ours & \textbf{7.54} & 17.96 & \textbf{9.22} & \textbf{19.49} & \textbf{29.97} & \textbf{15.82} & \textbf{24.58} & \textbf{13.15} & \textbf{17.22} \\
    \hline
    \end{tabular}
    \caption{The performance of various state-of-the-art algorithms trained via the Completion3D dataset on partial point cloud completion on a simulated partial point cloud dataset. 
    Results are reported on simulated partial point clouds of validation set.
    The Chamfer distance (CD) is reported, multiplied by $10^4$.}
    \label{table:completion3D_simulated}
\vspace{-3mm}
\end{table*}

\subsection{Ablation study}

Results of ablation study are shown in the Table \ref{table:ablation_study} and the overall classification accuracy is reported on the simulated partial test set of ModelNet40.
Unsurprisingly, the accuracy of models is improved after using the batchnorm~\cite{ioffe2015batch} and dropout~\cite{srivastava2014dropout} techniques, from 78.4\% to 81.0\%, shown by Model $A_i$.
We model the point clouds as an overlapping partition of point sets, each of which is defined by its centroid and scale.
As it is shown by Model $D_i$, the performance of this modeling is sensitive to the scale and accuracy peaks at 0.2 (86.4\%).
This can be in part explained by that local regions with small scale contain insufficient distinct geometry features to infer the complete point clouds, while the learned features for local regions with large scale tend to be different from those in partial point clouds with unexpected edges due to missing parts.
Model $E_i$ illustrates that the performance of model grows as we increase the dimension of the latent space (bottleneck), and it saturates at 2048 (86.8\%).

We propose to infer latent features voted from local point sets.
To make this voting strategy tenable, we design a training strategy that random number of votes are selected to compute the latent feature.
In the experiments, the maximum number of selected votes is manually set during training.
As it is shown by Model $B_i$, the performance peaks when the maximum number of votes is set to 10 (81.0\%) and degrades as more votes are considered.
We suspect that this is because more votes at training mean that more points are observed, and it leads to degradation at testing when only a small portion of partial objects are observed.
Model $C_i$ illustrates that the accuracy of the trained model grows as the number of votes increases at test from 76.8\% to 84.6\%, which indicates that more votes accumulate information for a more confidant prediction.

\subsection{Training Strategy Analysis}
We propose that a random number of votes are selected to compute the latent feature at training, which makes the voting strategy tenable and robust to the variant number of votes.
Since the latent feature is inferred from a subset of votes, the model at training makes predictions from partial shapes though complete shapes are taken as inputs.
Thus, the proposed training strategy can be seen as a way of data augmentation.
To make a fair comparison, we also report results of baseline models trained with the proposed training strategy, and they are shown in the column (Partial*) of the Table \ref{table:modelnet40}.
For other methods, the partial point clouds fed into the models at training consist of points covered by random sampled votes. 
Compared to results on partial shape while trained on complete shape in the Table \ref{table:modelnet40}, the proposed training strategy boosts the performance of PointNet (from 20.9\% to 76.5\%), PointNet++ (from 61.5\% to 81.9\%), RSCNN (from 43.3\% to 71.1\%), and DGCNN (from 61.5\% to 81.9\%).
However, our proposed method (86.4\%) still outperforms all of them with noticeable improvement, which verifies the effectiveness of our method.

\begin{figure}[t]
  \centering
  \includegraphics[width=0.8\linewidth]{{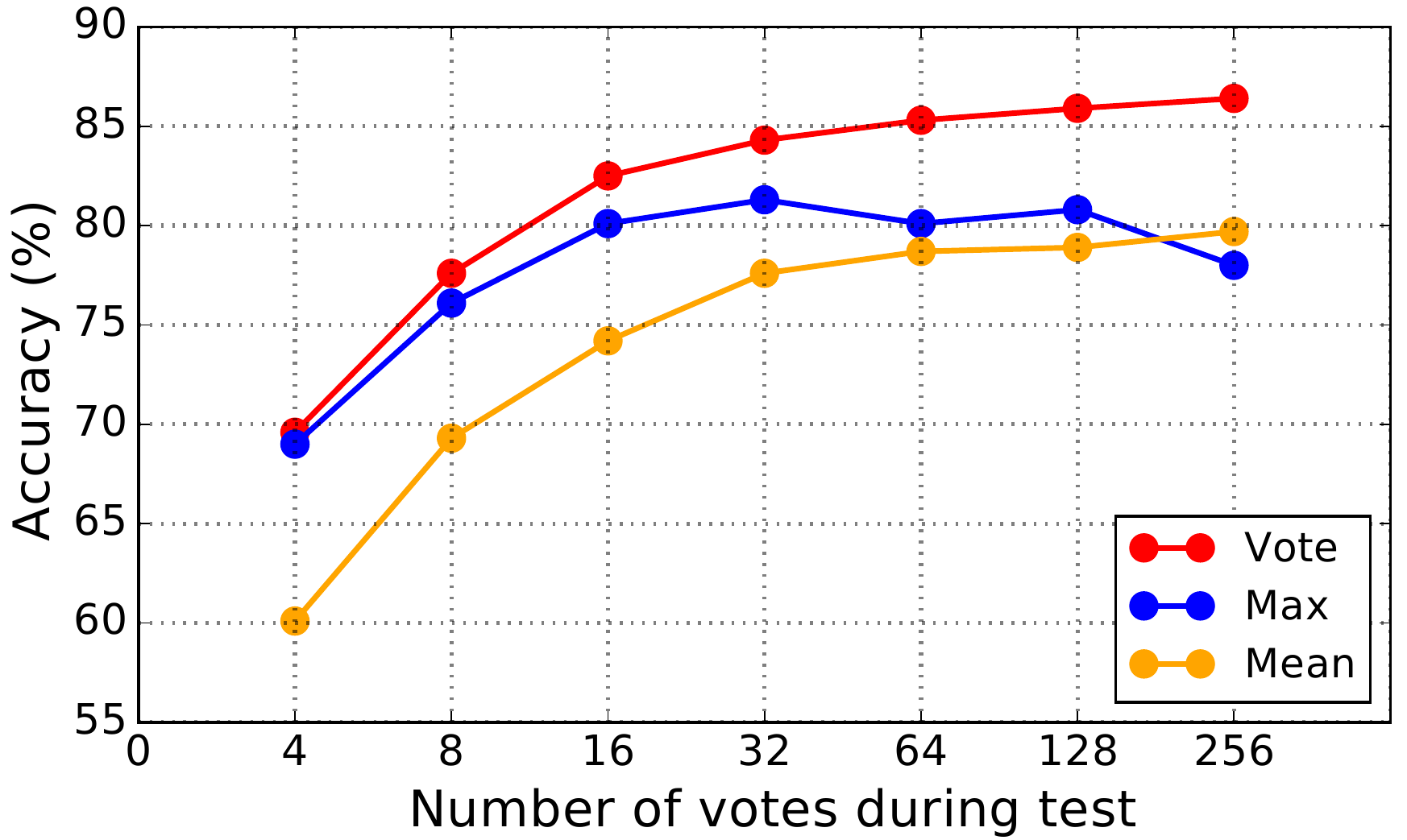}}
  \caption{Results of different aggregation strategies for computing latent features on simulated partial point clouds in ModelNet40. }
  \label{fig:aggre_analysis}
\end{figure}


\subsection{Voting Strategy Analysis}

As proposed in this work, we infer the latent feature from votes of local point sets, and each vote is a distribution in the latent space.
The optimal latent feature is the sampled one with the highest probability. 
In this section, we analyze this voting strategy and compare it with the aggregation strategy that the extracted features from local point sets are aggregated using a symmetric function. 
We study two symmetric functions: max pooling and mean pooling. 
To utilize the aggregation strategy, we change the vote of each local point set from a distribution to a deterministic feature vector.

The comparison results are shown in the Figure \ref{fig:aggre_analysis}. 
The evaluation metric is the average classification accuracy on the simulated partial ModelNet40. 
All models are trained using the proposed training strategy where the maximum number of votes considered is set to 10. 
Our proposed voting strategy is verified by the improved accuracy when compared to the aggregation strategy using either max pooling or mean pooling. 
Classification accuracy grows as the number of votes increases during testing in both the voting strategy and mean pooling.
This is because more votes accumulate more information for prediction. 
However, this is not the case for max pooling, whose accuracy peaks at 32 votes.
This indicates that max pooling is sensitive to the number of selected votes.

\subsection{Generalizability to different partial point clouds}
We experiment on different partial point clouds to verify the generalizability of the proposed model. 
In this section, we target at the point cloud completion task.
In addition to evaluating test set withheld in Completion3D dataset, which are shown in the Table \ref{table:completion3D}, we more exhaustively evaluate all approaches by experimenting on simulated partial point clouds as in the left subfigure of Figure \ref{fig:partial_point_clouds}.
Note that all models are trained on the provided training set in Completion3D dataset. 

\begin{figure}[t]
    \centering
    \includegraphics[width=\linewidth]{{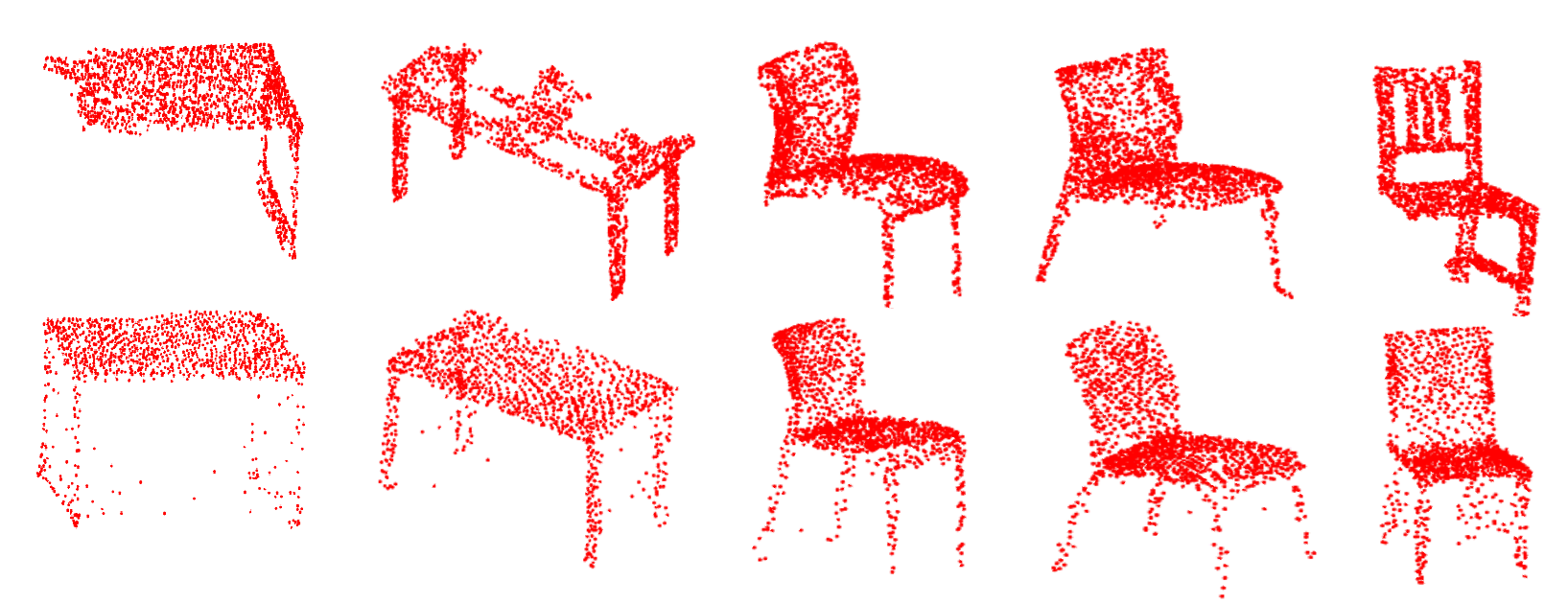}}
    \caption{Completion on real-world point clouds from ScanNet. Top row: input partial point clouds. Bottom row: complete point clouds generated by ours.}
    \label{fig:completion_scannet}
\vspace{-3mm}
\end{figure}

As shown in Table \ref{table:completion3D_simulated}, all approaches except the one developed in this paper experience a significant performance drop when compared to the results shown Table \ref{table:completion3D}.
We suspect that this is because the partial point clouds in the Completion3D are different from the simulated partial point clouds, as it is shown in the Figure \ref{fig:partial_point_clouds}.
Unlike other approaches embedding all input points into an encoding feature, we propose to rely on each local point set as the basic voter, which is more generalized and less affected by partial shapes.
This in part explains that our proposed method performs equally well on both experiments. 
Qualitative completion results are shown in Figure \ref{fig:completion_simulated}, and it shows that our proposed model outputs sharper shapes.  
More challenging tests are performed on real-world point clouds and are shown in the Figure \ref{fig:completion_scannet}. 
These partial point clouds are extracted within the labeled object bounding boxes provided in ScanNet~\cite{dai2017scannet}.
Note that the input point clouds need to be transformed to the box’s coordinates and the same scale as the training dataset before being fed into model.
This is because that the representations learned by the proposed model are not designed to be invariant to rigid body transformation.

\begin{figure}[t!]
    \centering
    \includegraphics[width=0.9\linewidth]{{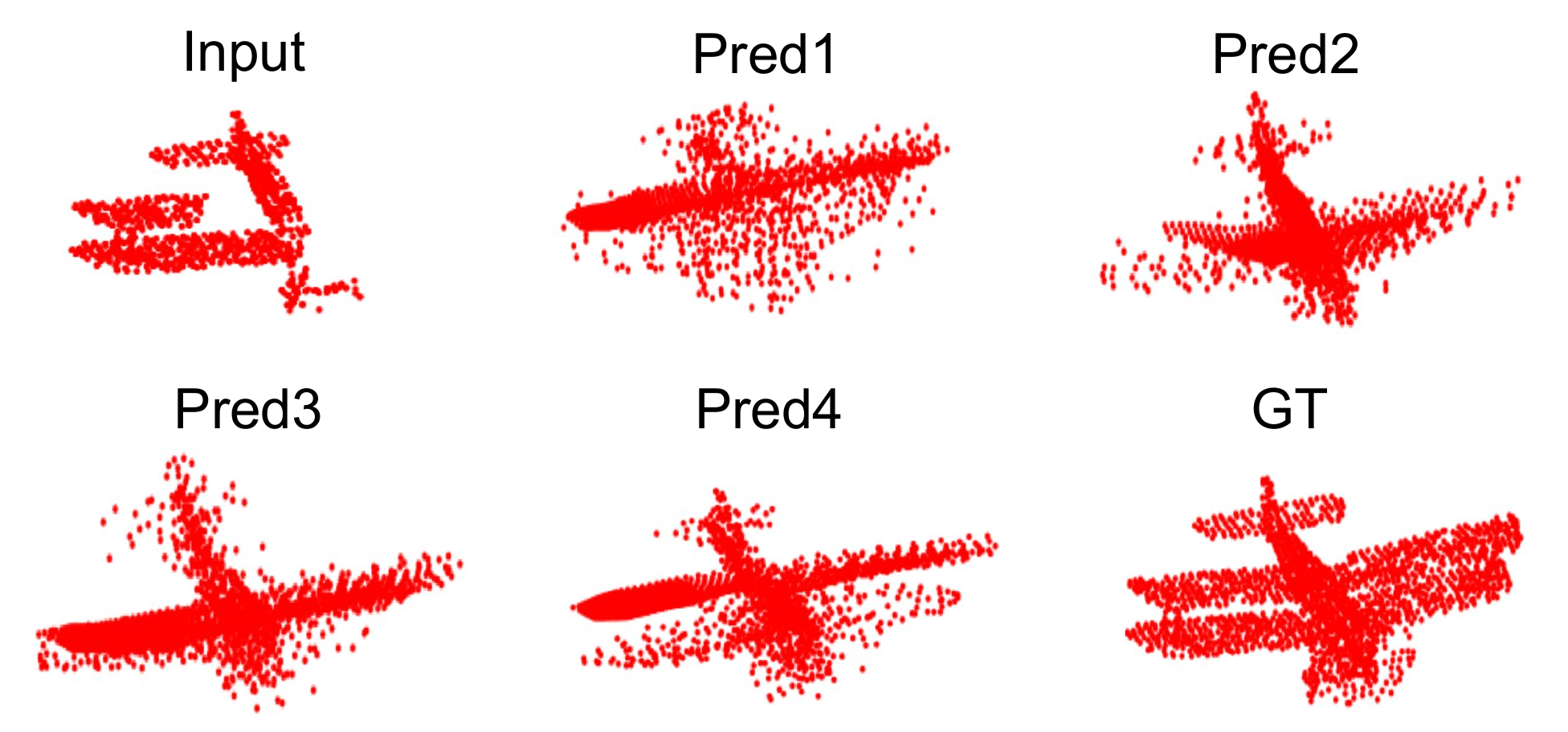}}
    \caption{Visualization of diverse predictions on point clouds completion on the Completion3D dataset.}
    \label{fig:multi_prediction}
\end{figure}

\subsection{Visualization of multiple predictions}

The method developed in this paper is designed to be able to generate multiple possible outputs.
This is achieved by the latent space model.
The latent space is represented by a set of multivariate Gaussian distributions generated by local point sets.
The use of latent value with the highest probability produces a single deterministic prediction. 
Diverse predictions can be generated by sampling from the latent space and followed by the decoding module.
Since it is not easy to sample in the latent space as it is represented by a set of distributions, we instead sample latent value by interpolating between the optimal latent feature inferred by all votes and a optimal latent feature inferred by a single vote. We perform experiments on point clouds completion and visualize the results in Figure \ref{fig:multi_prediction}.

\section{Conclusions}
This paper proposes a general model for partial point clouds analysis. 
In particular, point clouds are modeled as a partition of point sets which generate votes to model a latent space distribution. 
This voting strategy is shown to accumulate partial information and be robust to partial observation. 
The sampled latent feature in the latent space is then decoded for prediction. 
Extensive experiments are performed in classification, part segmentation, and completion and state-of-the-art results on all of them demonstrate the effectiveness of the proposed method.

\bibliographystyle{IEEEtran}
\bibliography{reference}

\end{document}